\documentclass[10pt,twocolumn,letterpaper]{article}

\usepackage[pagenumbers]{wacv} 

\usepackage{graphicx}
\usepackage{amsmath}
\usepackage{amssymb}
\usepackage{booktabs}
\usepackage[table]{xcolor} 
\usepackage{multirow}
\usepackage[accsupp]{axessibility} 
\usepackage{pifont} 

\usepackage[pagebackref,breaklinks,colorlinks]{hyperref}

\usepackage[capitalize]{cleveref}
\crefname{section}{Sec.}{Secs.}
\Crefname{section}{Section}{Sections}
\Crefname{table}{Table}{Tables}
\crefname{table}{Tab.}{Tabs.}

\def\wacvPaperID{24} 
\def\confName{WACV}
\def\confYear{2025}

\begin{document}

\title{MambaTron: Efficient Cross-Modal Point Cloud Enhancement using Aggregate Selective State Space Modeling}

\author{Sai Tarun Inaganti\\
Robotics Institute,\
University of Minnesota\\
{\tt\small inaga015@umn.edu}
\and
Gennady Petrenko\\
Homothereum\\
{\tt\small gennady@homothereum.org}
}
\maketitle

\begin{abstract}
Point cloud enhancement is the process of generating a high-quality point cloud from an incomplete input. This is done by filling in the missing details from a reference like the ground truth via regression, for example. In addition to unimodal image and point cloud reconstruction, we focus on the task of view-guided point cloud completion, where we gather the missing information from an image, which represents a view of the point cloud and use it to generate the output point cloud. With the recent research efforts surrounding state-space models, originally in natural language processing and now in 2D and 3D vision, Mamba has shown promising results as an efficient alternative to the self-attention mechanism. However, there is limited research towards employing Mamba for cross-attention between the image and the input point cloud, which is crucial in multi-modal problems. In this paper, we introduce MambaTron, a Mamba-Transformer cell that serves as a building block for our network which is capable of unimodal and cross-modal reconstruction which includes view-guided point cloud completion.We explore the benefits of Mamba's long-sequence efficiency coupled with the Transformer's excellent analytical capabilities through MambaTron.  This approach is one of the first attempts to implement a Mamba-based analogue of cross-attention, especially in computer vision. Our model demonstrates a degree of performance comparable to the current state-of-the-art techniques while using a fraction of the computation resources.
\end{abstract}

\section{Introduction}
\label{sec:intro}
In recent years, the point-cloud format has garnered attention in the 3D computer vision research community due to properties like packaging efficiency and applications in various domains like scene reconstruction and understanding, robotics, geographical information systems and autonomous driving. The point cloud is generally an unordered collection of data-points with each point represented by its 3D location coordinates (x, y, z). Its space complexity is only dependent on the number of points contained, unlike that of the other formats like volumetric grids for which it is cubic. Pointnet\cite{pointnet} pioneered a class of point cloud processing techinques that operated directly on the input instead of converting it into a voxel grid.

One practical problem with the raw point cloud data collected from 3D scanners like LiDAR is that they tend to be sparse, noisy and incomplete due to factors like occlusion, hardware limitations and optical phenomena like reflections and transparency. The data needs to be enhanced with density and completeness before it can be usable for downstream tasks like classification and segmentation.

Point cloud completion~\cite{surveypcenhance} is a category of enhancement where the missing data in an incomplete point cloud is recovered. In order to advance the research methodologies behind point cloud enhancement and completion, datasets like PCN~\cite{pcn}, KITTI~\cite{kitti}, ModelNet~\cite{modelnet}, ShapeNet~\cite{shapenet} and ShapeNet-ViPC\cite{vipc} were introduced, among which were CAD models or raw data directly collected from hardware like LiDAR and laser scanners. Metrics like Chamfer Distance~\cite{cd}, Earth Mover's Distance~\cite{emd, cd} and Jensen-Shannon Divergence\cite{jsd} were considered to measure the performance of point cloud completion techniques~\cite{surveypointcloud}. Parametrized completion techniques were introduced which primarily depended on the 3D prior knowledge of the input point cloud alone with no other auxiliary data. This approach was susceptible to accuracy errors, for example, in situations where two incomplete point clouds which belonged to different classes, but most of the difference is in the missing data that needs to be estimated. ViPC~\cite{vipc} introduced a multimodal processing technique where the missing information is inferred and enforced from a reference view or image which significantly aided in closing in the accuracy compared to unimodal methods.

\begin{figure*}[htbp]
    \centering
    \includegraphics[scale=0.95]{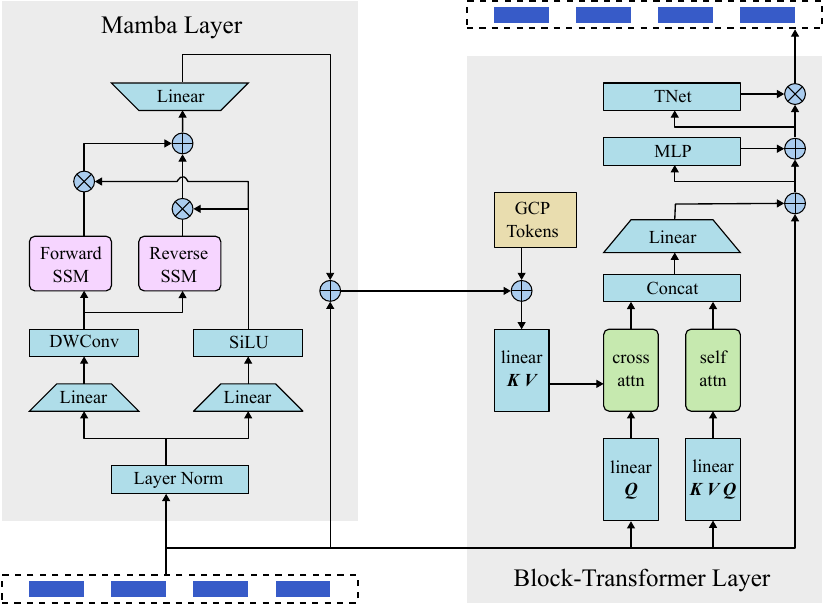}
    \caption{Illustration of the fully-bidirectional MambaTron cell. Input embeddings block is at the bottom-left in addition to the output block at the top right. The Mamba layer processes the entire input sequence at once while the Block-Transformer layer operates on a block of tokens at a time. We set the block size to 4 as illustrated above. The context state embeddings output by the Mamba layer are added to the Geometric Center Position (GCP) tokens which are the key-point coordinates while being fed to the Block-Transformer.}
    \label{fig:mambatron}
\end{figure*}

The transformer\cite{transformer} and recently the state-space model\cite{mamba, surveymamba360} architectures have contributed significantly to the field of natural language processing. They have been adapted for 2D~\cite{vit, vim} and 3D~\cite{pointbert, pointmae, pointmamba, pointramba, pcm} computer vision tasks with impressive results. Block-based transformers\cite{brect,block-transformer,bst} have tackled the problem of long token sequences by operating on blocks or windows of tokens at any given time. Our research adapts these natural language techniques towards the task of view-guided point cloud completion. In this paper, we detail our contributions, for which the summaries are given below.
\begin{enumerate}
    \item We propose the fully bidirectional MambaTron block, which consists of a Mamba State Space Model (SSM) layer for long-range global contextualization and an attention block for short-range neighborhood attention. This block is responsible for generating the piecewise contextualized local embeddings for both the auxiliary view and the partial point cloud.
    \item We present our end-to-end neural network model for point cloud reconstruction that optionally takes a reference view as input and use it to train the MambaTron cells. We use the MambaTron cell to encode the multimodal input (point cloud OR image) and also model the relationship between the two inputs (point cloud AND image).
    \item The Mamba layer is sensitive to the order in which the input tokens are presented and so we describe our Adjacency-Preserving Reordering (APR) technique to optimize MambaTron's performance while maintaining geometric adjacency among neighboring tokens.
    \item We demonstrate the performance on the ShapeNet-ViPC dataset, compare it with the SOTA image-assisted point cloud completion methods and show how our MambaTron-based network keeps up with state-of-the-art techniques with a fraction of resource utilization.
\end{enumerate}

\begin{figure*}[htbp]
    \centering
    \includegraphics[scale=0.95]{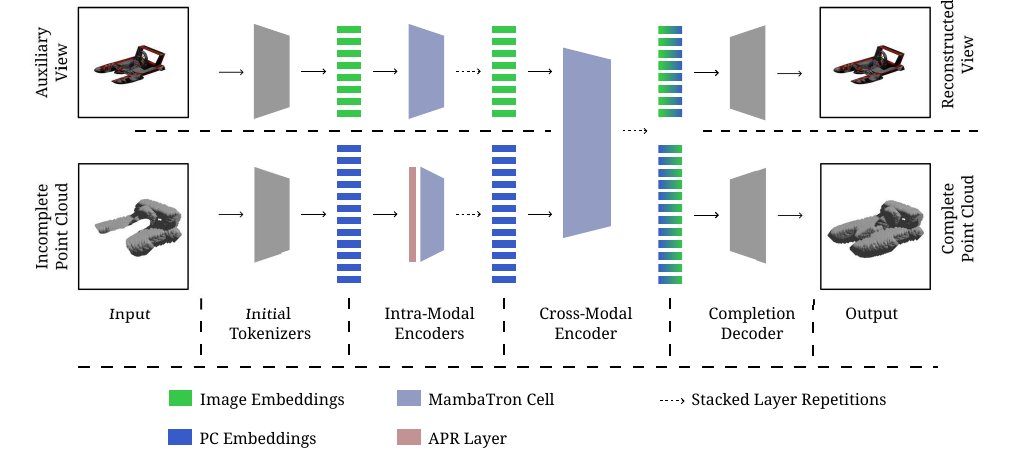}
    \caption{An overview of our MambaTron-based model for the View-Guided Point Cloud Completion Task.}
    \label{fig:model}
\end{figure*}

\section{Related Work}
\subsection{Point Cloud Analysis and Completion}
Studies were conducted on deep-learning parametric methods surrounding point cloud analysis~\cite{surveypointcloud} and enhancement~\cite{surveypcenhance, surveypccomplete} pioneered by PointNet's~\cite{pointnet} method to directly process 3D points and PCN's~\cite{pcn} ability to directly operate on point clouds without any assumptions about any of the global characteristics like structural information. PointNet++~\cite{pointnet++} was an encoder model that popularized the Farthest Point Sampling (FPS) method to downsample 3D points which helped the model learn local features which both the point-level and global features missed and promoted compactness in the input. FPS was adapted for its effectiveness by several other encoders~\cite{pointmlp, pcm, pointbert, pointmae, pointmamba, pointramba} and also in end-to-end completion techniques~\cite{snowflakenet, 3dmambacomplete}. Various architectural solutions were repurposed from the domain of natural language and speech processing like transformers~\cite{transformer, pointbert} and state space models\cite{mamba,pointmamba, pcm}. 2D vision also had some ideas to offer, like masked autoencoders~\cite{mae, pointmae}, GANs~\cite{gan} and DDPMs~\cite{ddpm, ddpmpc}.

On top of these ideas, which mostly described the fundamental structure and operations of the encoder/decoder layers, various point cloud enhancement techniques were recently introduced to tackle problems inherent to point clouds like surface properties, sparsity etc. GCN-based methods~\cite{pointgcn, dgcnn} use an encoder that relys on a graph that connects the input points as vertices, even in the feature space. FoldingNet~\cite{foldingnet} and AtlasNet~\cite{atlasnet} deform a 2D texture grid onto a 3D surface in a series of folding operations. TopNet~\cite{topnet} and SnowflakeNet~\cite{snowflakenet} implement specialized decoder modules to upsample/grow a point cloud in a tree pattern. Mamba-based enhancement methods~\cite{3dmambacomplete, 3dmambaipf} have leveraged Mamba's ability to efficiently process long 3D point sequences. Unimodal point cloud enhancement methods like the ones mentioned above have only the input priors to rely on and are more prone to performance degradation in completion tasks, compared to denoising and upsampling tasks, unless the input is constrained by class, geometry etc. These models are bound to be fully supervised as they rely on the availability of the ground truth point clouds.

\subsection{View-Guided Point Cloud Completion}
In order to combat the problem faced by point cloud completion tasks, the ViPC~\cite{vipc} architecture proposed using an auxiliary input like an image which was a view of the object in the input data that contained the missing information. This spanned the class of view-assisted methods~\cite{surveypccomplete, vipc, csdn, xmfnet, egiinet} that leveraged multimodal information fusion to generate a fulfilled point cloud. With the help of auxiliary views, the models can limit their reliance on the availability of ground truths by self-supervision. XMFnet~\cite{xmfnet} is a transformer-based model that utilizes cross-attention to fuse the multimodal features to form the completed output. EGIInet~\cite{egiinet} improves on top of XMFnet by adapting and incorporating style and content losses\cite{styletransfer} during the feature fusion process. These models are primarily attention based which have a quadratic computational complexity $O(L^2)$ with respect to the input token sequence length $L$. Our MambaTron-based completion network aims to reduce this to a near-linear theoritical complexity. Our network design is based on the cross-modal feature transfer mechanism of Joint-MAE~\cite{jointmae} and the point cloud groupwise-decoder designs of AXform~\cite{axform} and Joint-MAE.

\subsection{State Space Models}
\subsubsection{History}
\label{sec:ssm-history}
The Structured State Space Sequence (S4) model~\cite{s4} is the result of applying the principles of control theory in combination with sequential deep learning techniques like RNNs to model languages. It is the first attempt at efficiently processing long sequence data, surpassing its predecessor LSSL~\cite{lssl} by a wide margin in state computation efficiency, both of which utilize the HiPPO matrix~\cite{hippo} for state representation. These works spawned further SSM variants which made incremental progress towards modeling performance. Mamba~\cite{mamba} discards the linear time invariance (LTI) of its HiPPO matrices and introduces selection as a way to efficiently compress and package the context, which enabled SSMs to outperform transformer-based models while maintaining subquadratic computational complexity. 

Following its success in language modeling, Mamba was adapted for computer vision tasks. ViM~\cite{vim} adapts SSMs for image processing tasks as an analogue to the ViT~\cite{vit}. Point Mamba~\cite{pointmamba} orders point cloud patches using Hilbert/Trans-Hilbert ordering operations before feeding them to a Mamba layer for downstream tasks. PCM~\cite{pcm} stacks Mamba encoders and uses a Geometric Affine Module (GAM)~\cite{pointmlp}, uses code-based reordering and coupled encoder-decoder pairs for segmentation tasks. PoinTramba~\cite{pointramba} uses a transformer layer to generate patch features and a learning-based ordering technique to optimize Mamba's performance. 3dmambacomplete~\cite{3dmambacomplete} is an end-to-end point cloud completion technique that implements grid-folding in the decoder. 

Despite the advantages of SSMs, studies~\cite{hhh, illusionssm} have shown the limitations on the extent by which SSMs outperform transformers. MambaTron aims to overcome this by placing Mamba with a Block-Transformer to carry over the benefits of both the architectures while minimizing the bottlenecks.

\subsubsection{Analytical Study}
Quantitatively, S4~\cite{s4} introduces the Structured SSM which implements a linear control system with parameters $(\boldsymbol{\Delta}, \mathbf{A}, \mathbf{B}, \mathbf{C})$ in two stages. This system is a continuous sequence-to-sequence transformation $x(t) \longrightarrow y(t)$ that maintains a latent state $h(t)$ and is formulated below:

\begin{equation}
  h'(t) = \mathbf{A} h(t) + \mathbf{B} x(t), \quad y(t) = \mathbf{C} h(t)
  \label{eq:ssm-continuous}
\end{equation}

\noindent \textbf{Discretization.} The first stage implements the continuous-to-discrete parameter transformation $(\boldsymbol{\Delta}, \mathbf{A}, \mathbf{B}) \longrightarrow (\overline{\mathbf{A}}, \overline{\mathbf{B}})$ using a discretization rule $(f_A, f_B)$ like the zero-order hold (ZOH).

\begin{equation}
  \overline{\mathbf{A}} = f_A(\boldsymbol{\Delta}, \mathbf{A}), \overline{\mathbf{B}} = f_B(\boldsymbol{\Delta}, \mathbf{A}, \mathbf{B})
  \label{eq:discretization}
\end{equation}

\noindent \textbf{Computation.} The second stage computes the discrete sequence-to-sequence transformation $x_t \longrightarrow y_t$ with state $h_t$ using the new set of paramaters $(\overline{\mathbf{A}}, \overline{\mathbf{B}}, \mathbf{C})$.

\begin{equation}
  h_t = \overline{\mathbf{A}} h_{t-1} + \overline{\mathbf{B}} x_t, \quad y_t = \mathbf{C} h_t
  \label{eq:ssm-discrete}
\end{equation}

\noindent Here, $x_t \in \mathbb{R} $, $y_t \in \mathbb{R} $, $h_t \in \mathbb{R}^{N} $, $\overline{\mathbf{A}} \in \mathbb{R}^{N \times N} $, $\overline{\mathbf{B}} \in \mathbb{R}^{N \times 1} $, $\mathbf{A} \in \mathbb{R}^{N \times N} $, $\mathbf{B} \in \mathbb{R}^{N \times 1} $, $\mathbf{C} \in \mathbb{R}^{N \times 1} $ where $N$ is the length of the latent state vector $h$.

S4 models are Linear Time Invariant (LTI), meaning both the continuous parameters $(\boldsymbol{\Delta}, \mathbf{A}, \mathbf{B}, \mathbf{C})$ and discrete parameters $(\overline{\mathbf{A}}, \overline{\mathbf{B}})$ are fixed for all time steps. Mamba/S6 introduces the selective scanning mechanism and turns the parameters $\boldsymbol{\Delta}, \mathbf{B}, \mathbf{C}$ into time-variant functions of $x_t$ which have a length dimension $L$.

\subsection{Block-based Transformers}
 Language Transformers have mostly replaced general RNNs like LSTMs due to their parallel processing capabilities and attention mechanism which enables them to handle long input sequences without vanishing gradients. But their quadratic complexity quickly makes training transformer models on long input sequences expensive and unstable. To handle the complexity problem, various block/window-based architectures were proposed which limit the size of the attention matrix by introducing variations of state/context vectors. Block-Recurrent Transformer (BRecT)~\cite{brect} cells, inspired by LSTM cells, employ a sliding window to maintain a state vector like in an RNN while limiting a dimension of the attention matrix by the window size $W << L$, bringing the complexity down from $\mathcal{O}(L^2)$ of the vanilla transformer to $\mathcal{O}(LW)$. Block-transformers~\cite{block-transformer} operate on blocks of tokens, and employ a block-decoder followed by a token-level decoder, giving the same complexity. Block-State Transformers (BST)~\cite{bst} replace the recurrent cell in BRecTs with an SSM and processes each block in parallel instead of utilizing a sliding window, further bringing down the complexity to $\mathcal{O}(W^2) + \mathcal{O}(L \log L)$. The MambaTron cell is based on the BRecT cell and benefits from subquadratic complexity like BSTs.


\section{Qualitative Method Description}
\subsection{Problem Statement}
Our ultimate goal is to train a unified network that is capable of both unimodal reconstruction and cross-modal point cloud completion. We say reconstruction (reconstructed output is the same as the input) for unimodal inputs and not completion, which is beyond the scope of this work. We define multiple conditions that must be fulfilled in order to achieve this.
\begin{enumerate}
    \item If an incomplete point cloud and a complete reference image are given, the network generates embeddings that represent the complete image and also a complete version of the point cloud.
    \item If only a complete point cloud is given, the model must generate embeddings which can be decoded back to the input. This case is meant for downstream tasks and not enhancement.
\end{enumerate}

We design a model that utilizes MambaTron cells to generate embeddings for images and point clouds while incorporating missing information from the other modality when available. We later observe that the two conditions are complementary and help with the performance of the task corresponding to the other condition.

\subsection{MambaTron}
The word "MambaTron" is a portmanteau of three terms: Mamba, Transformer and Perceptron. The MambaTron cell takes advantage of the selective scanning mechanism of S6 and its expressive power to compute the context states and embeddings from the input sequence. It consists of two layers, one after the other. The cell and its inner workings are depicted in Figure \ref{fig:mambatron}.

\noindent \textbf{Mamba Layer.} Given an input sequence of length $L$, this layer processes the input bidirectionally so that every element in the input contributes to the context corresponding to a given input embedding. It outputs the context state vectors corresponding to each input embedding. The context sequence inherits its length $L$ from the input sequence.

\noindent \textbf{Block-Transformer Layer.} This layer divides the input sequence into blocks of size $W << L$ and processes them blockwise. For any given block, this layer's computation only depends on the the block's input and context, not on any other block's data, which means that they can be processed in parallel. This layer outputs the context-imbued feature embeddings of length $W$ per block. The output blocks are then concatenated back to give the output sequence of length L.

\subsection{Adjacency-Preserving Reordering}
\label{sec:apr}
In the point cloud encoding research works, there is a general consensus that the encoder must be able to capture three different types of embeddings to a fair extent: individual positional embeddings, the local neighborhood group embeddings and the global embeddings. Farthest Point Sampling (FPS)~\cite{pointnet++} to obtain a set of key points followed by a neighborhood selection algorithm like the K Nearest Neighbors (KNN) technique to divide the point cloud into groups of the same size, collectively abbreviated as FPS+KNN is a popular technique whose output serves as the input to the point cloud tokenizers that gives us the positional embeddings of the key points and also the collective embeddings of each neighborhood group. Depending on the tokenizer, pooling the positional embeddings can give us the global features. The feature embeddings of the point cloud groups are ordered before being sent to the Mamba encoder layer in any Mamba-based point cloud encoder.

\begin{figure}[ht]
    \centering
    \includegraphics[scale=0.80]{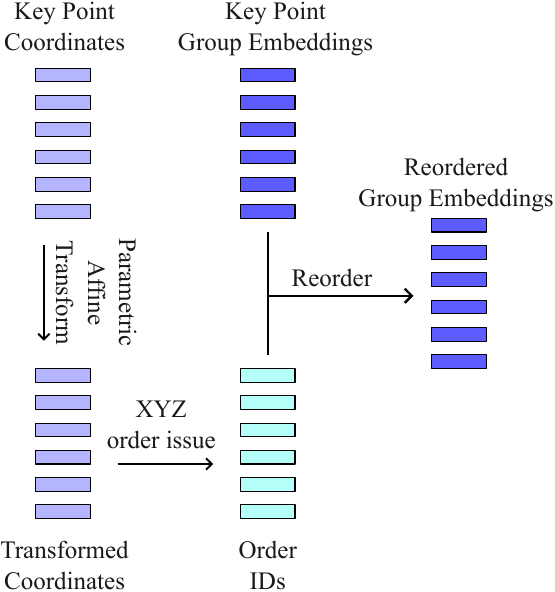}
    \caption{The Adjacency-Preserving Reordering (APR) scheme}
    \label{fig:apr}
\end{figure}

As witnessed in some of the previous works~\cite{pointmamba, pointramba, pcm} involving point cloud encoding tasks surrounding SSMs, Mamba's performance in computer vision tasks is greatly influenced by the ordering of the input tokens to the SSM layer. The previous works introduce ordering techniques to combat this which is described in Section \ref{sec:ssm-history}. The problem with these techniques is that they are either too rigid (like PointMamba's Hilbert ordering) and apply the same order for every layer or are too flexible (like PoinTramba's BIO ordering). PCM's code-based XYZ ordering technique allows for flexibility while maintaining a level of rigidity, i.e. adjacent groups in the input sequence are adjacent geometrically, which is what ViM~\cite{vim} goes for in 2D image encoding. However, PCM's technique is still too rigid as any point cloud is only privy to six different styles of ordering based on the permutations of the X, Y and Z directions. We argue that introducing more adjacency-respecting orders helps optimize Mamba's performance further. To achieve this, we introduce the Adjacency Preserving Reordering (APR) technique.

In the technique, we take the sparse point cloud containing the key points from the FPS algorithm and run it through a Transformation Network (TNet) which rotates and/or mirrors it while preserving the structure. Then we run the XYZ ordering scheme on the new point cloud. Since the number of possible affine transformations is infinite, there are an infinite ways to order the point cloud groups while maintaining the adjacency, only limited by the finite number of key points in the point cloud. Since a TNet is involved, this is a parameter-based ordering technique like BIO.

\begin{table*}[t]
\centering
\caption{Quantitative results on the original \textbf{ShapeNet-ViPC} dataset for the \textbf{Known} categories.}
\begin{tabular}{l|ccccccccc}
\toprule
Methods & Avg & Airplane & Cabinet & Car & Chair & Lamp & Sofa & Table & Watercraft \\
\midrule
\multicolumn{10}{c}{\textit{Mean CD per point $\times 10^{-3}$ (lower is better)}} \\
\midrule
ViPC~\cite{vipc}    & 3.308 & 1.760 & 4.558 & 3.183 & 2.476 & 2.867 & 4.481 & 4.990 & 2.197   \\
CSDN~\cite{csdn}    & 2.570 & 1.251 & 3.670 & 2.977 & 2.835 & 2.554 & 3.240 & 2.575 & 1.742   \\
XMFnet~\cite{xmfnet}    &  1.443 & 0.572 & 1.980 & 1.754 & 1.403 & 1.810 & 1.702 & 1.386 & 0.945   \\
EGIInet~\cite{egiinet} & 1.211 & \textbf{0.534} & 1.921 & 1.655 & 1.204 & 0.776 & 1.552 & \textbf{1.227} & 0.802 \\
\rowcolor{gray!20} 
Ours & 1.199 & 0.537   & 1.905   & 1.620   & 1.184 & 0.774   & 1.561   & 1.234   & 0.779   \\
\midrule
\multicolumn{10}{c}{\textit{Mean F-Score @ 0.001 (higher is better)}} \\
\midrule
ViPC~\cite{vipc}    & 0.591 & 0.803 & 0.451 & 0.512 & 0.529 & 0.706 & 0.434 & 0.594 & 0.730   \\
CSDN~\cite{csdn}    & 0.695 & 0.862 & 0.548 & 0.560 & 0.669 & 0.761 & 0.557 & 0.729 & 0.782   \\
XMFnet~\cite{xmfnet}    &  0.796 & 0.961 & 0.662 & 0.691 & 0.809 & 0.792 & 0.723 & 0.830 & 0.901   \\
EGIInet~\cite{egiinet} & 0.836 & \textbf{0.969} & 0.693 & 0.723 & 0.847 & 0.919 & 0.756 & \textbf{0.857} & 0.927 \\
\rowcolor{gray!20} 
Ours & 0.839 & 0.969   & 0.697   & 0.728   & 0.851 & 0.920   & 0.764   & 0.856   & 0.931   \\
\bottomrule
\end{tabular}
\label{tab:shapenet-vipc-known}
\end{table*}

\subsection{Model}
\noindent Our model pipeline is a series of four sections, as shown in Figure \ref{fig:model}.
\begin{enumerate}
    \item Initial Tokenizer
    \item Intra-modal Encoder
    \item Cross-modal Encoder (for cross-modal instances only)
    \item Reconstruction Decoder
\end{enumerate}

\noindent In this paper, we focus on the design of sections 2 and 3 of which the MambaTron cell is the building block.


\section{Quantitative Model Execution}

We are presented with a point cloud $(N, 3)$ with or without an auxiliary image $(H, W)$, the end-to-end neural network model from Figure \ref{fig:model} receives as input and outputs a reconstructed version of the input point cloud.

\subsection{Forward Pass}

\subsubsection{Initial tokenizer}
We borrow PointMamba's technique to extract the initial tokens which gives us the key point cloud $(n_p, 3)$ and the corresponding position embeddings $(n_p, C)$ and group embeddings $(n_p, C)$. We reference ViM for the 2D counterpart, which gives us the image patch embeddings $(n_i, C)$.

\subsubsection{Intra-Modal Encoder}
This is a serially stacked MambaTron encoder which takes in either the image or the point cloud tokens that are output by the tokenizer and output their feature embeddings imbued with the global context. For the image input, we prepend a special $<I>$ token and append a $<STOP>$ token. For the point cloud input, we first reorder the tokens using APR from section \ref{sec:apr}, then replace $<I>$ with $<P>$ along with adding the GCP tokens to the context vectors as shown in Figure \ref{fig:mambatron}, which are basically the key point positional embeddings. The output feature embeddings $(n_i, C)$ or $(n_p, C)$ with the special tokens removed are imbued with global context by this encoder.

\subsubsection{Cross-Modal Encoder}
This encoder is used to extract context from one modality and impart it to the other. The image and point cloud group embeddings are arranged in the following order to form the input.
\begin{enumerate}
    \item $<I>$ token
    \item image embeddings
    \item $<P>$ token
    \item key point group embeddings
    \item $<STOP>$ token
\end{enumerate}

\noindent Similar to the intra-modal encoder, the group context vectors are supplied with the GCP tokens at the block-transformer layer. The outputs $(n_i, C)$ and $(n_p, C)$ with the special tokens removed now contain the relationships of each image patch or point cloud group with every token from the other modality.

\subsubsection{Decoder}
The decoder takes the embeddings $(n_i, C)$ or $(n_p, C)$ from the intra-modal or the cross-modal encoder, depending on the number of inputs and reconstructs the image or the point cloud with the same dimension as the input. Due to the modularity of our method, the decoder is selected based on its ability to process image embeddings if needed and is task-specific which will be described later in section \ref{sec:experiments}.

\begin{table*}[t]
\centering
\caption{Quantitative results on the \textbf{ShapeNet-ViPC} dataset for the \textbf{Novel} categories.}
\begin{tabular}{l|cc|cc|cc|cc|cc}
\toprule
\multirow{2}{*}{Methods} & \multicolumn{2}{c|}{Avg} & \multicolumn{2}{c|}{Bench} & \multicolumn{2}{c|}{Monitor} & \multicolumn{2}{c|}{Speaker} & \multicolumn{2}{c}{Phone}\\
& CD & F & CD & F & CD & F & CD & F & CD & F \\
\midrule
ViPC~\cite{vipc}    & 4.601 & 0.498 & 3.091 & 0.654 & 4.419 & 0.491 & 7.674 & 0.313 & 3.219 & 0.535   \\
CSDN~\cite{csdn}    & 3.656 & 0.631 & 1.834 & 0.798 & 4.115 & 0.598 & 5.690 & 0.485 & 2.985 & 0.644   \\
XMFnet~\cite{xmfnet}    &  2.671 & 0.710 & 1.278 & 0.862 & 2.806 & 0.677 & 4.823 & 0.556 & 1.779 & 0.748   \\
EGIInet~\cite{egiinet} & 2.354 & 0.750 & 1.047 & 0.902 & 2.513 & 0.716 & 4.282 & 0.591 & 1.575 & 0.792 \\
\rowcolor{gray!20} 
Ours & 2.333 & 0.761   & 1.041   & 0.909   & 2.499 & 0.723   & 4.268   & 0.599   & 1.566 & 0.797 \\
\bottomrule
\end{tabular}
\label{tab:shapenet-vipc-novel}
\end{table*}

\subsection{Training}

\noindent Our training method is a two-stage process.

\noindent \textbf{Unimodal stage.} The model is trained on complete point cloud inputs. Even though this is called the unimodal training stage, we actually derive the top-down image projection of the input point cloud similar to Joint-MAE~\cite{jointmae} and pass it along as well, making this a cross-modal task. We also partially mask the inputs before the decoder stage to promote robustness. This stage teaches the model to develop an idea of how complete point clouds look like and how they correlate to images.

\noindent \textbf{Cross-modal stage.} After the first stage, the model is further trained with an incomplete point cloud and an auxiliary view as inputs and a completed point cloud as the output which is compared to the ground truth point cloud for regression.

We first look at the various types of training loss that we consider for regression in both the stages.

\noindent \textbf{Chamfer Distance.} This loss consolidates the reconstructed point cloud to the ground truth.
\begin{equation}
  \mathcal{L}_{CD} = \sum_{x \in P_1} \min_{y \in P_2} \| x - y \|_2^2 
+ \sum_{y \in P_2} \min_{x \in P_1} \| y - x \|_2^2
  \label{eq:loss-cd}
\end{equation}

\noindent \textbf{Style Loss.} The style loss ~\cite{styletransfer, adain, csdn, egiinet} regression  ensures that the model is learning to apply the image features to the point cloud and vice-versa. 
The style loss is defined as:
\begin{equation}
    \mathcal{L}_\text{style} = \frac{\left( G(F_{i}) - G(F_{p}') \right)^2 + \left( G(F_{p}) - G(F_{i}') \right)^2}{n_P \times C}
    \label{eq:loss-style}
\end{equation}
where $F_p$ and $F_i$ are the intra-modal features, whereas $F_p'$ and $F_i'$ are the cross-modal features. 
$G$ is the gram matrix function $G(F) = F^T \cdot F$.

\noindent \textbf{Projection Loss.} Similar to Joint-MAE's \cite{jointmae} reconstruction loss, we project the reconstructed point cloud onto a 2D plane by the same view of the input projection and compare it with the initial projection (unimodal stage only).

\begin{equation}
  \mathcal{L}_{proj} = MSE(I_{proj}(P_{output}), I)
  \label{eq:loss-proj}
\end{equation}

\noindent \textbf{2D Loss.} This loss compares the final image to the auxiliary view in the cross-modal case and the input projection.
\begin{equation}
  \mathcal{L}_{2D} = MSE(I_{output}, I)
  \label{eq:loss-2D}
\end{equation}

\noindent For the unimodal training stage, the training loss is below.
\begin{equation}
  \mathcal{L}_{uni} = \mathcal{L}_{CD} + \mathcal{L}_{2D} + \mathcal{L}_{proj}
  \label{eq:loss-uni}
\end{equation}

\noindent For the cross-modal stage, regression is performed using the following loss.
\begin{equation}
  \mathcal{L}_{cross} = \mathcal{L}_{CD} + \mathcal{L}_{2D} + \mathcal{L}_{style}
  \label{eq:loss-cross}
\end{equation}


\section{Experiments and Results}
\label{sec:experiments}
\subsection{Cross-Modal Point Cloud Completion}
We first train our model with a Joint-MAE~\cite{jointmae} based decoder at the unimodal stage using complete point clouds from the ShapeNet55 dataset which consists of images belonging to one of 55 categories, then train on the ShapeNet-ViPC dataset with an 80-20 train-test split which contains objects with noisy and noiseless views from 13 categories at the cross-modal stage. We set MambaTron's window size to 4. For the evaluation, we use the l2-normalized CD and the F-score metrics similar to the other state-of-the-art techniques for this task. On average, our model outperforms the latest state-of-the-art technique with only 3.92 M parameters compared to EGIInet's 9.03 M and XMFnet's 9.57 M. Our model captures the objects with more complex features and limited samples more effectively compared to EGIInet, especially on novel categories. The results are presented in Tables \ref{tab:shapenet-vipc-known} and \ref{tab:shapenet-vipc-novel}.

\begin{table}[t]
\centering
\caption{ScanObjectNN dataset, accuracy in \%}
\begin{tabular}{lccc}
\toprule
\multirow{2}{*}{Method} & Params. & FLOPs & PB\_T50\_RS \\
& (M) & (G) & $\uparrow$ \\
\midrule
Point-MAE~\cite{pointmae}   & 22.1 & 4.8 & 85.18 \\
Joint-MAE~\cite{jointmae}    &  - & - & 86.07  \\
PointMamba~\cite{pointmamba}    & 12.3 & 3.1 & 89.31 \\
PCM~\cite{pcm}    &  34.2 & 45.0 & 88.10  \\
\rowcolor{gray!20} 
Ours & 13.6 & 3.9   & 90.17     \\
RECON~\cite{recon} & 43.6 & 5.3 & 90.63  \\
\bottomrule
\end{tabular}
\label{tab:scanobjectnn}
\end{table}

\begin{figure}[ht]
    \centering
    \includegraphics[scale=0.80]{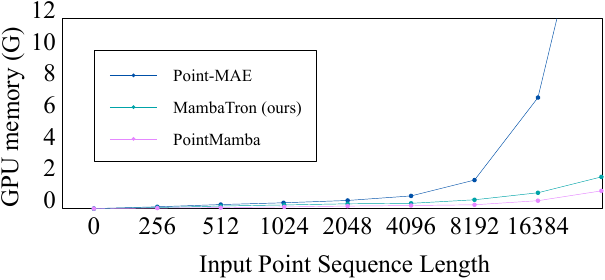}
    \caption{GPU usage comparision}
    \label{fig:plot-gpu}
\end{figure}

\subsection{Unimodal Point Cloud Analysis}
Though the main focus of this paper is view-guided point cloud completion, we also test the efficacy of our model on point cloud analysis tasks by pretraining. We train our model with an AXform-based~\cite{axform} decoder at the unimodal stage similar to the previous task on the ScanObjectNN~\cite{scanobjectnn} dataset, but this time with an 80-20 split. On the same 80\% training data, we replace the decoder with a classification head and train a second time. The window size is the same as before. We observe comparable results comparable to the SOTA methods on average while on a much smaller computational footprint, giving the accuracies of $94.97\pm0.02\%$ on OBJ\_BG, $93.15\pm0.02\%$ on OBJ\_ONLY and $90.17\pm0.02\%$ on PB\_T50\_RS, which are the evaluation metrics used.

We also test the computational efficiency \cite{pointmamba} of our MambaTron-based model where we map the GPU usage to the length of the input sequence to show how MambaTron's subquadratic complexity is superior to that of the transformer-based models, which is shown on Figure \ref{fig:plot-gpu}.

\subsection{Ablation Studies}
We study the effects of various design decisions like the cell level components, network level components, the multiple stages of training and the losses.

\noindent \textbf{Effect of the Block-Transformer.} A primary question would be the purpose of the Block-Transformer in the MambaTron cell. The Mamba layer of the MambaTron block focuses on the relationships among the input elements and not the elements themselves. But our design allows for smoothly swapping out the block, in which case the output embeddings are replaced by the context state vectors. We tested the model with the block-transformer disabled and found that the model performed worse than even the baseline ViPC model on the ShapeNet-ViPC dataset at its size with the novel categories being affected the most. For the pretraining task, the performance dropped below that of PointMamba. The MambaTron cell converged quicker than its bare Mamba counterpart, yet the performance still tanked on the novel categories.

\noindent \textbf{Effect of sharing the intra-modal encoder.} In our model, a common stacked MambaTron encoder accepts tokens from either the image or the point cloud and adds context to each token. Using a seperate intra-modal encoder for each encoder increased the number of parameters while the slowing down the training as the loss function takes longer to converge. The model performed worse on more complex and novel categories.

\noindent \textbf{Effect of the cross-modal encoder.} Since the shared intra-modal encoder already processes both the image and the point cloud, we tested the model without a cross-modal encoder or the style loss. The projection loss in the unimodal training stage stays intact.

\noindent \textbf{Effect of the style and projection losses.} This time, we trained the model without the mentioned losses contributing to the regression, while the cross-modal encoder is intact, to show the effect of the image on the encoder.


The ablation performance is recorded and the average CD for the ShapeNet-ViPC dataset are given in Table~\ref{tab:ablation} for the known and novel categories.

\begin{table}[t]
\centering
\caption{Ablation study on ShapeNet-ViPC}
\begin{tabular}{lcc}
\toprule
\multirow{2}{*}{w/o} & \multicolumn{2}{c}{CD (avg)} \\
     & known & novel \\
\midrule
-   & 1.199 & 2.333 \\
block-tr    &  1.754 & 3.217  \\
shared intra    & 1.317 & 2.625 \\
cross-modal    &  1.363 & 2.699 \\
style/proj loss & 1.322 & 2.676    \\
\bottomrule
\end{tabular}
\label{tab:ablation}
\end{table}

\section{Conclusion and Future Work}

In this paper, we have described the MambaTron cell and built a modular network model using cell for the tasks of view-guided point cloud completion and unimodal point cloud pretraining which can be used for downstream tasks like classification. Our model performs on par with or surpasses the current state-of-the-art models for both the tasks with a limited footprint while being quicker, owing to MambaTron's fast inference coupled with expressive power. This version of the MambaTron is limited to outputting embeddings which are constrained to the input's dimensions. Further research can remove this constraint, thus helping generalize the cell's applications. At the network level, improvements can be applied to further integrate the MambaTron cell into the model instead of simply utilizing it as a building block for a conventional feed-forward network. Future works can focus on using the cell for unguided point cloud completion in addition to other enhancement tasks like upsampling and denoising. We hope that our work helps other researchers in point cloud and multimodality research with applications like scene-awareness and path-planning in fields like Robotics.


\noindent \textbf{Acknowledgment.} This work was supported by the Minnesota Robotics Institute (MnRI).

{\small

}

\end{document}


\title{--- Supplementary Material ---\\MambaTron: Efficient Cross-Modal Point Cloud Enhancement using Aggregate Selective State Space Modeling}

\author{Sai Tarun Inaganti\\
Robotics Institute,\
University of Minnesota\\
{\tt\small inaga015@umn.edu}
\and
Gennady Petrenko\\
Homothereum\\
{\tt\small gennady@homothereum.org}
}
\maketitle

\section{Effects of the APR scheme}
The Adjacency Preserving Reordering (APR) scheme is space-filling, similar to the Hilbert~\cite{pointmamba} and XYZ~\cite{pcm} ordering techniques. Additionally, the scheme theoretically provides an infinite ways to order the points, only limited by the number of points, thanks to a learnable affine transformation layer. We look at the effects of each component through a further ablation study on the ShapeNet-ViPC dataset in Table~\ref{tab:ablation-apr}.

\begin{table}[t]
\centering
\caption{Effects of APR}
\begin{tabular}{lccc}
\toprule
\multirow{2}{*}{APR} & \multirow{2}{*}{affine} & \multicolumn{2}{c}{CD (avg)} \\
   &  & known & novel \\
\midrule
\xmark & \xmark & 1.354 & 2.737 \\
\cmark  & \xmark &  1.317 & 2.611  \\
\rowcolor{gray!20} 
\cmark  & \cmark &  1.199 & 2.333  \\
\bottomrule
\end{tabular}
\label{tab:ablation-apr}
\end{table}

\section{Performance statistics on additional datasets}
The task of cross-modal point cloud completion is a recent idea introduced~\cite{vipc} along with the ShapeNet-ViPC dataset, with no equivalent dataset that is just as comprehensive, i.e. contains class-mapped pointclouds with corresponding reference view images. For unimodal analysis, in addition to the real-world 3D object classification on the ScanObjectNN dataset, we perform experiments that we describe below with the same encoder configuration as that of the classification task on ScanObjectNN. The training data is subject to random rotation and slight scaling operations at each epoch. For comparision purposes, we take the results reported in the other research works.

\noindent \textbf{Classification on ModelNet40.} This~\cite{modelnet} is a synthetic CAD dataset with 40 classes that contains a total of 12,311 noiseless pre-aligned models. The classification head is based on that of PointMamba~\cite{pointmamba}. We report the Overall Accuracy (OA) in percentage in Table~\ref{tab:modelnet40}.

\begin{table}[t]
\centering
\caption{ModelNet40 dataset}
\begin{tabular}{lcc}
\toprule
Method & OA (\%)\\
\midrule
PCM~\cite{pcm} & 93.4 \\
PointMamba~\cite{pointmamba} &  93.6  \\
Point-MAE~\cite{pointmae} & 93.8 \\
Joint-MAE~\cite{jointmae} & 94.0 \\
\rowcolor{gray!20} 
Ours &  94.3 \\
\bottomrule
\end{tabular}
\label{tab:modelnet40}
\end{table}

\noindent \textbf{Segmentation on ShapeNetPart.} We perform the task of part segmentation on the ShapeNetPart~\cite{shapenetpart} dataset which contains 16,881 shapes from 16 categories, with 50 segmentation parts in total. We report the mean Intersection over Union (mIoU) percentages at both the class level and the instance level in Table~\ref{tab:shapenetpart}. Our model demonstrates superior performance by a good margin, especially on the instance level.

\begin{table}[t]
\centering
\caption{ShapeNetPart dataset}
\begin{tabular}{lcc}
\toprule
\multirow{2}{*}{Method} & \multicolumn{2}{c}{mIoU (\%)} \\
 & class & instance \\
\midrule
Point-MAE~\cite{pointmae} & 84.2 & 86.1 \\
Joint-MAE~\cite{jointmae} & 85.4 & 86.3 \\
PointMamba~\cite{pointmamba} &  84.4 & 86.2  \\
PCM~\cite{pcm} & 87.0 & 85.3 \\
\rowcolor{gray!20} 
Ours &  87.4 & 86.9  \\
\bottomrule
\end{tabular}
\label{tab:shapenetpart}
\end{table}


{\small

}